\documentclass{article}

\usepackage[final]{nips_2017}

\usepackage[utf8]{inputenc} 
\usepackage[T1]{fontenc}    
\usepackage{hyperref}       
\usepackage{url}            
\usepackage{booktabs}       
\usepackage{amsfonts}       
\usepackage{nicefrac}       
\usepackage{microtype}      

\usepackage[usenames, dvipsnames]{color}
\usepackage{amsmath,amssymb}
\usepackage{graphicx}
\usepackage[inline]{enumitem}
\usepackage{pifont}
\usepackage{subcaption}

\title{The Atari Grand Challenge Dataset}

\author{
  Vitaly Kurin
  \\
  Visual Computing Institute\\
  RWTH Aachen University\\
  Aachen, Germany\\
  \texttt{vitaliykurin@gmail.com} \\
  \And 
  Sebastian Nowozin, Katja Hofmann \\
  Machine Intelligence and Perception Group \\
  Microsoft Research \\
  Cambridge, UK \\
  \texttt{\{Sebastian.Nowozin, Katja.Hofmann\}@microsoft.com} \\
  \And
  Lucas Beyer, Bastian Leibe  \\
  Visual Computing Institute \\
  RWTH Aachen University \\
  Aachen, Germany \\
  \texttt{\{beyer, leibe\}@vision.rwth-aachen.de}
}

\DeclareMathOperator*{\argmax}{argmax}

\begin{document}

\maketitle

\begin{abstract}
 
Recent progress in Reinforcement Learning (RL), fueled by its combination, with Deep Learning has enabled impressive results in learning to interact with complex virtual environments, yet real-world applications of RL are still scarce. A key limitation is data efficiency, with current state-of-the-art approaches requiring millions of training samples.
A promising way to tackle this problem is to augment RL with learning from human demonstrations.
However, human demonstration data is not yet readily available. This hinders progress in this direction.
The present work addresses this problem as follows. We
\begin{enumerate*}[label=(\roman*)]
  \item collect and describe a large dataset of human Atari 2600 replays -- the largest and most diverse such data set publicly released to date
  \item illustrate an example use of this dataset
by analyzing the relation between demonstration quality and imitation learning performance, and
  \item outline possible research directions that are opened up by our work.
\end{enumerate*}
\end{abstract}

\section{Introduction}

In Reinforcement Learning (RL), an agent learns, by trial and error, to perform a task in an initially unknown environment. Recently, this research area has seen dramatic progress in complex interactive tasks in virtual environments \cite{mnih2015human,DBLP:journals/corr/MnihBMGLHSK16,silver2016mastering,DBLP:journals/corr/SchulmanLMJA15}, largely driven by combinations of RL with deep learning. Yet, despite this recent progress, real-world applications are still largely lacking.

In the RL setup, agents learn to solve a task completely from scratch. 
This causes one of the key limitations of state-of-the-art deep RL approaches --- data inefficiency.
In comparison to autonomous agents, humans have a lot of prior information about the world.
Every day, we base our decisions on knowledge about culture and social relationships, our own experience, and information we get from experience of others. 
As a result, when learning a new task, people are more effective and need less actions to master it.
During training, the agent executes a lot of actions that a human never would.
This is ineffective and renders the application of RL in complex or potentially dangerous environments infeasible. 
A possible solution to the problem is to learn from human demonstrations~ \cite{schaal1997learning,ng2000algorithms,abbeel2004apprenticeship,monfort2017asynchronous,hester2017learning}.

Up until now, the RL community has been focused on building environments --- test beds for RL models. 
Today, RL approaches can be trained and compared on diverse tasks in environments such as ALE~\cite{bellemare13arcade}, OpenAI Gym~\cite{1606.01540} or Microsoft's Project Malmo~\cite{johnson2016malmo}. 
But there are few publicly available datasets of human demonstrations of tasks in these environments. 
This lack hampers the progress of research on learning from human demonstration.
Examples such as ImageNet~\cite{deng2009imagenet} in Computer Vision and Switchboard~\cite{godfrey1992switchboard} in Speech Recognition have shown that datasets can catalyze research progress.
In order to accelerate research in learning from demonstration, we release, describe, and illustrate the use of the Atari Grand Challenge dataset\footnote{\texttt{http://atarigrandchallenge.com/data}} --- a dataset of human Atari 2600 replays.

Our contributions are:
\begin{enumerate*}[label=(\roman*)]
  \item We collect, analyze, and release to the research community the largest and most diverse dataset of human Atari 2600 replays to date. The dataset comprises ${\sim}9.7$ million frames (${\sim}45$ hours) of game play for five games - an order of magnitude larger than previous datasets.
  \item We illustrate one use of this dataset by analyzing the relation between demonstration quality and imitation learning performance.
  \item We discuss research directions that are opened up by our work.
\end{enumerate*}

\section{Background}

This section outlines key concepts and notation used throughout the paper. 
We operate in a usual Reinforcement Learning (RL) setup: an agent acts in an environment in response to state observations, and learns from a reward signal that reflects an abstract notion of consequences of the actions taken. 
This setup can be formulated as Markov Decision Process (MDP), defined by a tuple $\langle S,A,R,T \rangle$, where $S$ is the set of states, $A$ is the set of actions, $R(s,a)$ is the reward function, and $T(s,a,s')$ is the transition function that returns a probability over states, given a state and an action: $p(s'|s,a)$, $s,s' \in S$, $a \in A$. 
In each iteration of its interaction with the environment, the agent observes the state, takes an action and gets some reward for the transition. An agent's behavior is characterized by a policy $\pi(s)$, a function that returns an action given a state. The policy can be stochastic.

Our experimental analysis (section ~\ref{sec:data-quality}), 
is based on a recently proposed imitation algorithm suggested in~\cite{hester2017learning}, in turn based on Q-learning~\cite{watkins1992q} and in particular DDQN~\cite{van2016deep}. We next give an overview of the relevant approaches.

Q-learning~\cite{watkins1992q} is a ``model-free'' RL algorithm. It centers on learning to approximate the so-called action-value function $Q(s,a)$. The Q-function reflects the expected discounted cumulative value of taking a particular action $a$ in state $s$, and following the particular policy thereafter. 
The \textit{optimal} action-value function $Q^{*}(s,a)$ should satisfy the Bellman equation:

\begin{equation}
    Q^*(s, a) = \mathop{\mathbb{E}_{s'}}{\big[R(s, a) + \gamma \max_{a' \in A} Q^*(s', a') | s, a \big]}, 
\end{equation}

where $\gamma$ is a discount factor that trades off immediate versus longer term rewards.
The optimal policy $\pi^*(s)$ is the policy which takes the best possible decision on each time step:

\begin{equation}
    \pi^*(s) = \argmax_{a \in A} Q^*(s,a).
\end{equation}

DQN~\cite{mnih2015human} is a variant of Q-learning that uses a neural network (called Deep Q-Network) to approximate Q-values. 
The network $Q(s,\cdot;\theta)$ returns action values for all the actions available given the current state. A separate target network is used to compute Q-values in training updates, as well as a so-called replay memory of past experience for minibatch sampling. Both were shown to improve training stability and resulted in breakthrough results when learning to play Atari games.
Double DQN \cite{van2016deep} is an extension of DQN which decouples the selection and value estimate of actions in the max operator, which was shown to result in more accurate Q-value approximations both theoretically and in practice.
The learning objective looks as follows:

\begin{equation}
J_{DQ}(Q) = \big[R(s,a) + \gamma Q(s',a'_{max};\theta') - Q(s,a;\theta)\big]^2,
\end{equation}

where $a'_{max} = \argmax_{a' \in A} Q(s',a';\theta)$.

Recent work \cite{hester2017learning} suggests an approach to imitation learning that combines the Double DQN objective and a large margin classification loss aimed at keeping a learned policy close to demonstrated behavior:

\begin{equation}
    J(Q) = J_{DQ}(Q) + \lambda_1J_E(Q) + \lambda_2J_{L_2}(Q),
\end{equation}

where $J_{L_2}(Q)$ is the $L_2$ regularization,  $J_E(Q)$ is the supervised learning loss:

\begin{equation}
    \label{eq:hester-loss}
    J_E(Q) = \max_{a\in A}\big[Q(s,a) + l(s,a_E,a)\big] - Q(s,a_E),
\end{equation}

and  $l(s,a_E,a)$ is the large margin classification loss that returns some positive number if the expert's action $a_E \neq a$ and zero otherwise. 
The large margin classification loss prevents the learner from over-estimating Q-values for previously unseen states.
The Q-values for actions taken by an expert are forced to be a margin higher than those of the unseen ones.

All the methods described above can also be applied completely off-line on data collected in the process of a human's interaction with the environment. 

\section{Constructing the Atari Grand Challenge dataset}

This section details our approach to collecting the Atari Grand Challenge dataset. All described tools are made public with the data set.

\subsection{Collecting the dataset}

We collected our dataset using a web application built around Javatari\footnote{\texttt{http://javatari.org/}}, an Atari 2600 emulator written in JavaScript. 
Given an initial state and the full sequence of human inputs, the emulator is completely deterministic, and we can avoid the excessive burden of saving screenshots of the game at every single time step while it is being played. Instead, we can only record the initial state and player inputs, and generate the dataset of images offline by playback. 
This makes data collection at large scale feasible with limited resources.

After processing, for each time step we have a screenshot of the game, the action taken at that time step, the reward, the current score, and the information about the time step being terminal or not.
Since incomplete (non-terminal) episodes still carry useful information, we save the episode each time a player closes the application tab or browser window as well as when the game ends.

All the Atari-related functionality is entirely processed on the client side --- within a browser. 
We support all the major web browsers: Google Chrome, Mozilla Firefox, Microsoft Edge, Safari.
The server is built using Flask\footnote{\texttt{http://flask.pocoo.org/}}.
It is only responsible for saving the data and, later, for loading the data when replaying. 
All the data is saved in a PostgreSQL database.
The replay process is automated with Selenium\footnote{\texttt{https://github.com/SeleniumHQ}}.

Our case is a good example of gamified crowdsoursing --- using people's desire to play to do useful things. In order to engage people more, we added two progress bars: one compares players' performance with the best human player result, the other shows the same comparison with DQN performance taken from~\cite{mnih2015human}.

\subsection{Dataset post-processing}

There are two steps of dataset post-processing. First, we try to eliminate the differences between the Javatari emulator and the ALE. 
The only difference we have found is that the states we get in Javatari are vertically shifted by several pixels in comparison to ALE states. 
We eliminate this difference by shifting the states as in ALE and padding them with zeroes at the bottom and top borders.

During the first frames that we record, the emulated Atari memory is not fully initialized and we might get an excessively large score, which is not correct. We therefore fix the first several frames' rewards to zero. 
We are not also interested in games where the person did not interact with the application more than just opening and closing it. 
We filter these cases by simply removing games with a final score of zero.
The post-processing is fully automated, and new data can be processed with the code provided.

\section{Properties of the Atari Grand Challenge dataset}
\label{sec:stats}

\subsection{Description of the dataset}

In this section we briefly describe what the Atari Grand Challenge dataset consists of and we show some of its properties that we deem particularly relevant to research on learning from human demonstrations.
 
\subsubsection{Scale}

The dataset consists of human replays for five popular Atari 2600 games: Video Pinball, Q*bert, Space Invaders,  Ms.Pacman  and Montezuma's Revenge. 
The choice of the games is not random: we want to vary the level of difficulty according to the results in~\cite{mnih2015human}. 
The DQN was able to play the first game significantly better than human players, the results for the second and the third were comparable to human performance and the latter two were very hard for DQN.

\begin{figure}[ht]
    \begin{center}
        \centerline{\includegraphics[width=\columnwidth]{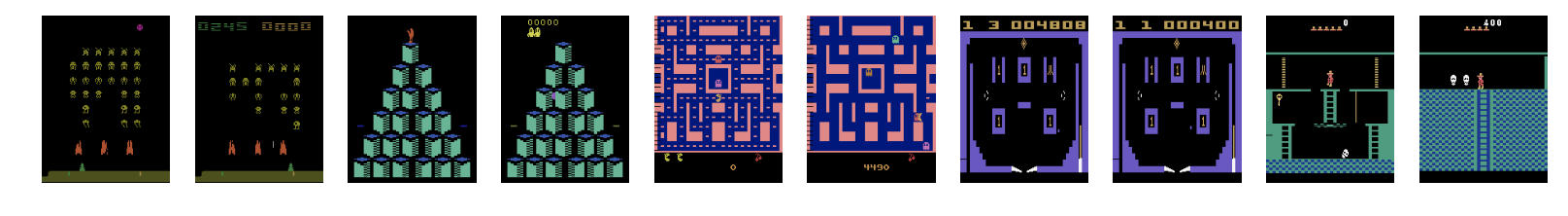}}
        \caption{Sample screenshots from the dataset (from left to right, the games shown are: Space Invaders, Q*bert, Ms. Pacman, Video Pinball, Montezuma's Revenge)}
        \label{fig:samples}
    \end{center}
    \vskip -0.2in
\end{figure}

The Atari Grand Challenge dataset consists of 2367 game episodes with positive final score, that is ${\sim}9.7$ million frames or ${\sim}45$ hours of playing time at 60 frames per second. 
Table~\ref{table:dataset-stats} shows per-game statistics of the dataset and Fig.~\ref{fig:samples} shows sample screenshots.

\begin{table}[t]
\caption{Atari Grand Challenge per-game statistics}
\label{table:dataset-stats}
\centering
\resizebox{.99\textwidth}{!}{
\begin{tabular}{llllll} 
\toprule
&Space Invaders&Q*bert& Ms.Pacman& Video Pinball& Montezuma's revenge\\
\midrule
episodes&445 &659 & 384 &211 &668 \\ 
frames &2,056,741 &1,599,453 &1,779,771&1,526,215 &2,717,676\\ 
gameplay, hrs &9.52 &7.40 &8.23& 7.05&12.57\\ 
worst score & 5&25 &10&100 &100\\ 
best score &3,355 & 41,425&29,311&67,150&27,900\\ 
\bottomrule
\end{tabular}
}
\end{table}

\subsubsection{Diversity}
\label{subsec:diversity}



\begin{figure}[ht]
    \begin{center}
        \centerline{\includegraphics[width=\columnwidth]{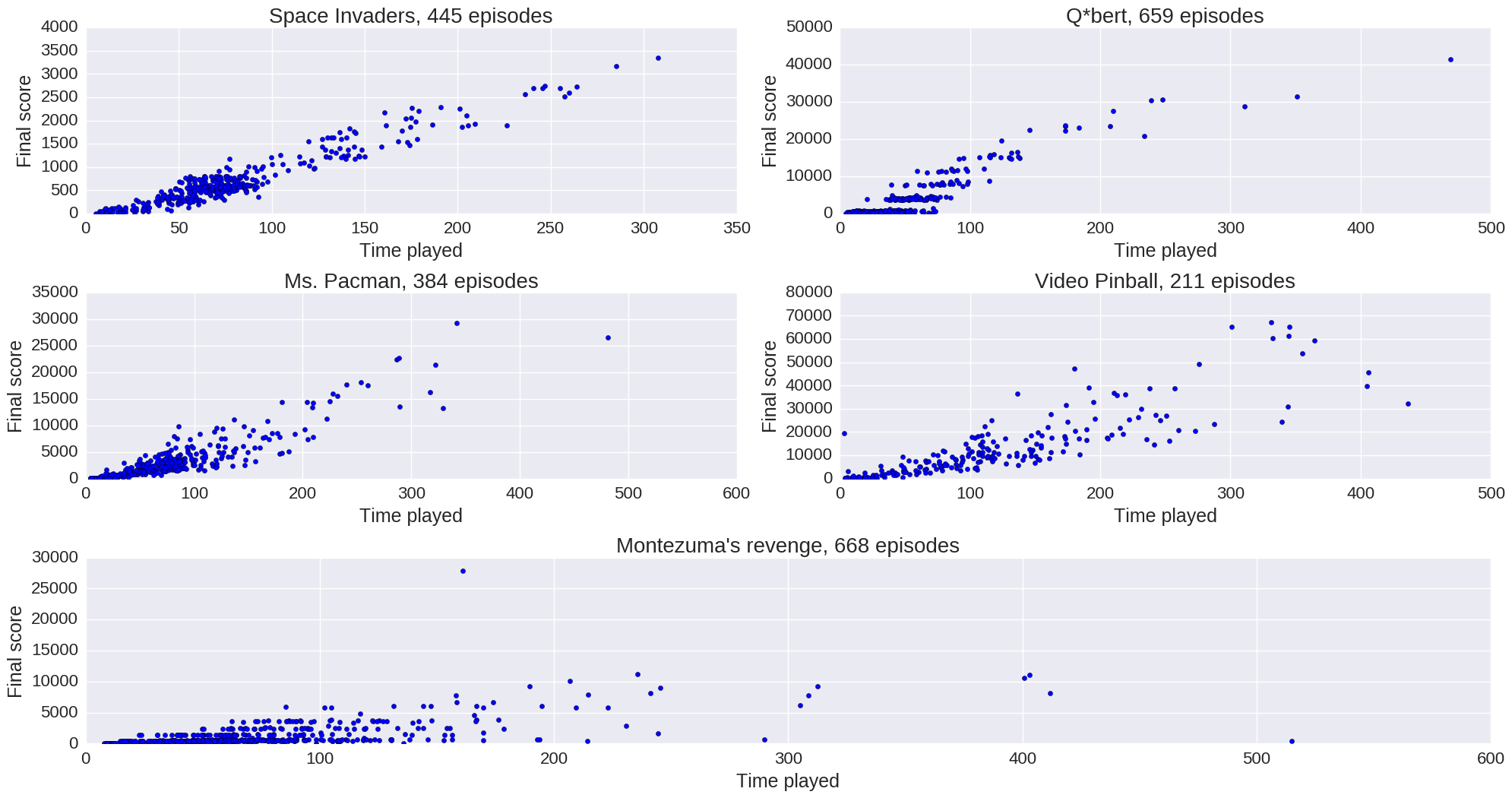}}
        \caption{Atari Grand Challenge dataset human demonstrators score dependency on time.}
        \label{fig:time-vs-score}
    \end{center}
    \vskip -0.2in
\end{figure} 

Since the data is collected in the wild, some of the players were good, some of them were bad. 
As a result, Fig.~\ref{fig:time-vs-score} shows that the Atari Grand Challenge dataset is quite diverse in terms of the final score distribution. From the episode final score and the time played, we can already make some assumptions about the different players' level of expertise.
What else can we do to show the player diversity quantitatively?

\begin{figure}[ht]
    \begin{center}
        \centerline{\includegraphics[width=\columnwidth]{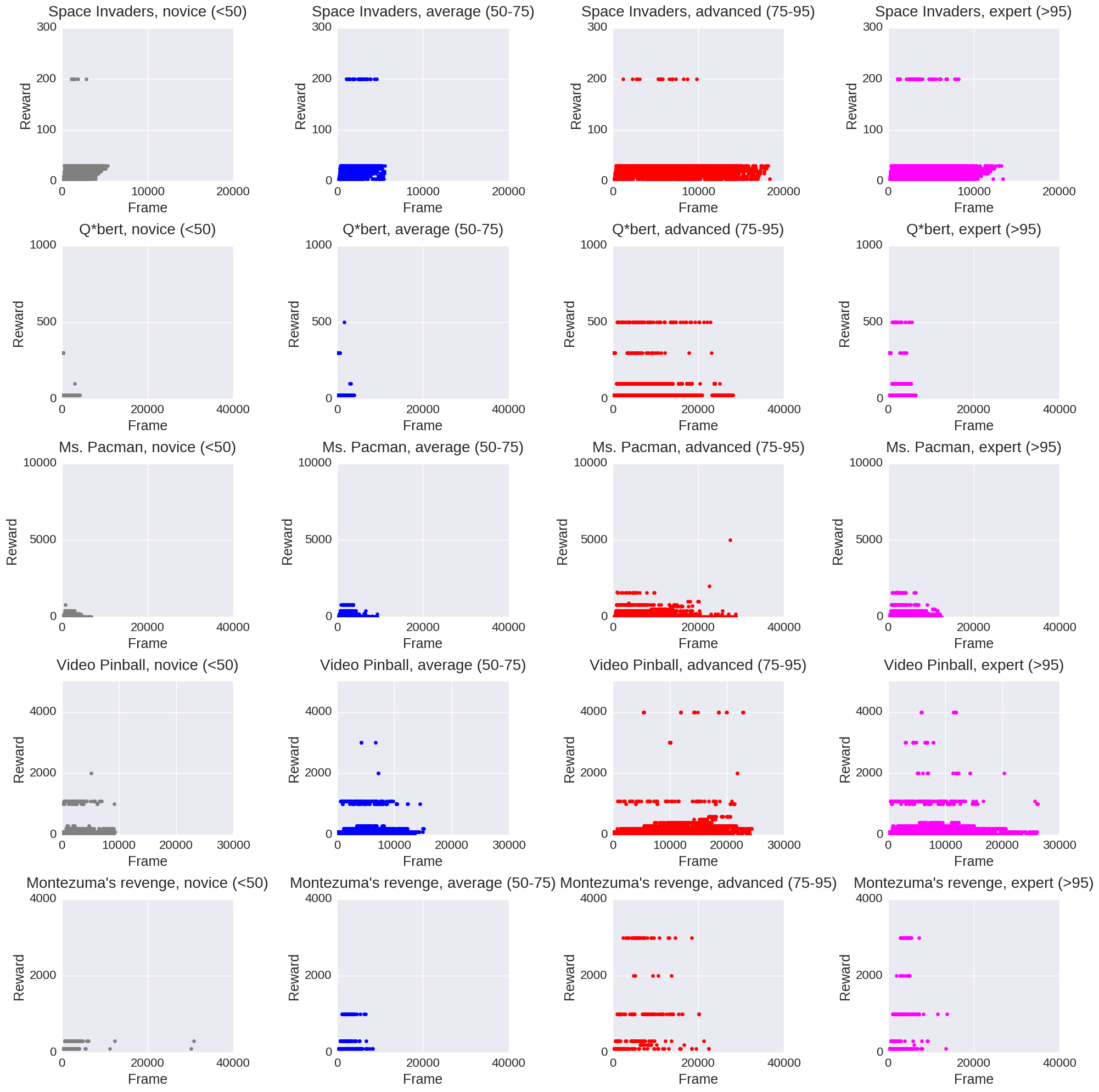}}
        \caption{Reward and the frame when the reward was obtained show that all the players have equal access to rewards, but more experienced players are more efficient. All the game episodes are divided as follows: ``novice'' (below 50 percentile of all the scores), ``average'' (between 50 and 75), ``advanced'' (between 75 and 95) and ``expert'' (above 95 percentile).}
        \label{fig:reward-availability}
    \end{center}
    \vskip -0.2in
\end{figure}

It is natural to assume that the more experienced a player is, the more effective he will be. 
A new player cannot achieve a challenging reward when an experienced player can. 
Fig.~\ref{fig:reward-availability} shows that all the players have equal access to the rewards (at least for the games in question).
From the comparison of ``advanced'' and ``expert'' groups we can see, that ``expert'' players are faster in achieving the rewards: the rightmost column data points look more shifted to the left. 
Given that the final score for the ``advanced'' group is higher, they achieve more in shorter periods of time.

\subsubsection{Extensibility}

Currently the dataset comprises five Atari 2600 games, but it can be easily extended by adding a new game. 
We do not only publish the dataset, but the code for data collection as well\footnote{\texttt{https://github.com/yobibyte/atarigrandchallenge/}}.
Any of the Atari 2600 non-paddle games available in ALE \cite{bellemare13arcade} can be added within few hours of work.
We do not support paddle games like Breakout or Pong since it is almost impossible to exactly repeat the noisy controller to collect the states off-line.

\section{Influence of data quality on the imitation learning performance}
\label{sec:data-quality}

\subsection{Experiments description}
\label{subsec:experiments}

As shown in Section~\ref{sec:stats}, we have replays of good players as well as replays of bad players. 
In this section we show how the dataset can be used to study how the demonstrator expertise can influence the performance of imitation learning.

In this experiment, we filter the training data by a minimum score. We train our model on the frames of the episodes with final score above a threshold: 50 percentile, 75 percentile (top 25\% of the data) and 95 percentile (top 5\%). 
We also train the model on the whole dataset. 

We train the model completely off-line as in Eq.~\ref{eq:hester-loss}, but we do not use the regularization term from~\cite{hester2017learning} since we have more data to train on.
We use $\lambda_1=1.0$ and $l=0.8$ for non-expert actions as suggested in the same paper.
We use the same network architecture as in~\cite{mnih2015human} and train over $10^6$ iterations.
We use the Adam~\cite{kingma2014adam} optimizer with learning rate $\epsilon=0.00025$ and $\beta_1=\beta_2=0.95$.
The target network update interval is $10,000$, mini-batch size is 32, and training is run for one million updates.
The code is written in Chainer~\cite{chainer_learningsys2015} within the Malmopy\footnote{\texttt{https://github.com/Microsoft/malmo-challenge}} framework.
Since we do not have any frameskip during data collection, we use frameskip coefficient $k=1$. We normalize reward values by dividing the raw rewards by the largest reward value observed in our data. 
There were no negative rewards in our case.
The experimentation code can be found on our github page\footnote{\texttt{https://github.com/yobibyte/atarigrandchallenge/}}.

\subsection{Results}

\begin{table}[!t]
\caption{Average score $\pm$ standard error of the mean (out of 100 games) for models trained on subsets of the data filtered by a score. The first four rows use the offline part of the imitation algorithm (without regularization) from~\cite{hester2017learning} to train on our data. Top 5\% means that the training data consists of the episodes with a final score higher or equal than the 95 percentile score. Evaluation is performed with the $\epsilon$-greedy policy, $\epsilon = 0.05$. \cite{mnih2015human}~reports the standard deviation of the scores, but we do not have information about the number of the episodes and thus we cannot report their SEM.}
\label{tab:results-filtering}
\centering
\resizebox{.99\textwidth}{!}{
\begin{tabular}{llllll} 
\toprule
&\multicolumn{1}{p{0.5cm}}{\centering Space \\ Invaders}&Q*bert& Ms.Pacman&\multicolumn{1}{p{0.3cm}}{\centering Video \\ Pinball}&\multicolumn{1}{p{0.5cm}}{\centering Montezuma's \\revenge}\\
\midrule
Imitation All data &$125\pm9.94$&$146\pm14.87$&$250\pm18.09$& 8,823$\pm$745.26 &$7\pm4.32$\\ 
Imitation top 50\% &$90\pm8.64$&$127\pm13.80$&$308\pm20.66$& 11,216$\pm$801.53 &$4\pm1.97$\\ 
Imitation top 25\% &$127\pm9.69$&$179\pm17.01$&$271\pm22.15$& 24,351$\pm$2,084.38 &$22\pm8.11$\\ 
Imitation top 5\%&$144\pm12.40$&$545\pm107.19$&$418\pm19.98$& 17,775$\pm$16.10 &$36\pm7.98$\\
Imitation~\cite{hester2017learning} & n/a & 5,133.8&$692.4$&10,655.5&576.3\\
\midrule
DQN~\cite{mnih2015human}& 1,976 & 10,596 & 2,311 & 42,684 & 0\\
DDQN~\cite{van2016deep}& 2,628.7 & 11,020.8 & 1,241.3 & 367,823.7 & 42 \\
Random uniform &156$\pm$9.46& 162$\pm$17.6 &211$\pm$11.35 & 30,368$\pm$3,723.87 & 0$\pm$0 \\
\bottomrule
\end{tabular}
}
\end{table}

During training we evaluate the performance of the model on 100 episodes every 100,000 mini-batch updates. 
After training we take the model with the best average results and re-evaluate it on 100 games and report the average score and standard error of the mean.
Table~\ref{tab:results-filtering} is in line with our hypothesis: the higher the filter value for the data, the better the performance.
The sub-par performance of the imitation model on our dataset in three out of five games can be explained by looking at Table~\ref{tab:hester-vs-atari}: our data has lower and more diverse human demonstrator scores than those of~\cite{hester2017learning}.
At the same time, in Video Pinball, where our data has better human scores, the model performs better.


\begin{table}[t]
\centering
\caption{Comparison of human scores in \cite{hester2017learning} and the Atari Grand Challenge dataset.}
\label{tab:hester-vs-atari}
\resizebox{.99\textwidth}{!}{
\begin{tabular}{lllllll}
\toprule
                    & \multicolumn{3}{c}{\cite{hester2017learning}}               & \multicolumn{3}{c}{Atari Grand Challenge dataset}                 \\
\cmidrule{2-7}
                    & Worst score & Best score & \#transitions & Worst score & Best score & \#transitions \\
\midrule
Space Invaders      & \multicolumn{3}{c}{n/a}                  & 5           & 3355       & 2,056,741     \\
Q*bert              & 80700       & 99450      & 75472         & 25          & 41425      & 1,599,453     \\
Ms. Pacman          & 31781       & 55021      & 21896         & 10          & 29311      & 1,779,771     \\
Video Pinball       & 8409        & 32420      & 10051         & 100         & 67150      & 1,526,215     \\
Montezuma's revenge & 32300       & 34900      & 17949         & 100         & 27900      & 2,717,676     \\
\bottomrule
\end{tabular}
}
\end{table}
\section{Related work}

There are two directions of RL research which are working on leveraging demonstration data for training an autonomous agent: Inverse Reinforcement Learning (IRL) and Imitation Learning. 
The former group addresses scenarios where there is no access to the reward function. 
It is true that in RL tasks the goal is often underspecified, and sometimes it is hard to provide a reward that represents all the useful information from expert's demonstration. 
The general idea is to approximate the reward function and learn a policy using this approximation~\cite{ng2000algorithms,abbeel2004apprenticeship}.

Whilst IRL can benefit from the Atari Grand Challenge dataset by ignoring the reward information, Imitation Learning is the direct benefactor of our dataset.
Imitation learning exploits the reward information to learn an action-value function, or directly a policy.
\cite{schaal1997learning} uses a pre-trained model to speed up training, and has an interesting comparison of pre-training influence on model-free and model-based RL.
The paper notes that model-based learning benefits more from using demonstration data.
The latest work on Learning from Demonstration shows that model-free RL can also greatly benefit from using human player data~\cite{hester2017learning,subramanian2016exploration,hosu2016playing}.

The datasets collected for the learning from demonstration research described above are either small, or not available for public use.
The Atari Grand Challenge dataset is the largest and the most diverse in terms of the types of the games as well as amount and types of human players release so far.

Up until now, the RL community has been mostly focusing on building the environments for training autonomous agents: ALE~\cite{bellemare13arcade}, OpenAI's gym~\cite{1606.01540} and Universe\footnote{\texttt{https://github.com/openai/universe}}, Microsoft's Malmo project~\cite{johnson2016malmo}. 
The final goal of operating in these environments is in maximizing the final score. 
Even if we train our models off-line, it sounds reasonable to evaluate the performance within such an environment. 
That is why, in Table~\ref{tab:dataset-comparison} below we describe the datasets coupled with interactive environments.



Atari 2600 games have recently begun to take a similar role as experimentation ground for RL research as MNIST has taken for computer vision and many implementations of RL algorithms have been evaluated on such games.
Therefore, it is much easier to compare leveraging the human behavior data with pure RL implementations or even combining them.

\begin{table}[t]
\caption{Learning from demonstration datasets comparison. The replay data for~\cite{hosu2016playing} has not been published, but there are Montezuma's Revenge and Private Eye checkpoints --- saved states of the environment that can be used for continuing the episode.}
\label{tab:dataset-comparison}
\centering
\resizebox{.99\textwidth}{!}{
\begin{tabular}{llllcc} 
\toprule
&Domain&Tasks&Size (transitions)&Open&\multicolumn{1}{p{2.2cm}}{\centering Diverse in \\player expertise}\\
\midrule
Atari Grand Challenge
&Atari 2600&5&${\sim}9.7$ mil.&\ding{51}&\ding{51}\\
Udacity self-driving dataset\footnotemark&
Driving simulator
&1&8086 (x3 cameras)&\ding{51}&\ding{55}\\
\cite{hosu2016playing}&Atari 2600&1&${\sim}1.2$ mil.&(\ding{55})&\ding{55}\\
\cite{hester2017learning}&Atari 2600&42&${\sim}1$ mil.&\ding{55}&\ding{55}\\
\bottomrule
\end{tabular}
}
\end{table}

\section{Discussion and Future work}

Our work opens up a wide range of follow-up work on benefits and uses of human demonstrations for effectively and efficiently learning to interact with complex environments.

\subsection{Extending the Atari Grand Challenge dataset}

The Atari Grand Challenge website is still on-line and people keep playing. 
We plan to update the dataset in the future as more data becomes available. 
The most important development of the dataset is to collect more data of ``professional players'' who achieve higher scores. 
As we have shown, the data quality affects the final performance dramatically, it will be a good improvement, when we do that. 

\subsection{Exploiting the Atari Grand Challenge dataset}

Video games are a perfect testing ground for evaluating hypotheses and learning how we can use human data to achieve higher sample efficiency and make the RL training process faster. 
So, our future research will focus on improving sample efficiency of RL algorithms by leveraging the data of diverse quality.

In this paper we have shown just one of the possible dataset applications: how data quality influences the final performance of imitation learning~\cite{hester2017learning}. 
We hope that researchers in machine learning, game AI and maybe even cognitive science, can find something useful for their own research purposes. We find the following applications particularly appealing.

Recently, Inverse Reinforcement Learning and Imitation Learning have regained popularity~\cite{ho2016generative,baram2016model,hester2017learning}. 
Our dataset has a direct impact on this kind of research.
It is interesting to check if can we take something useful out of bad players data.
Even experienced players make mistakes.
But throwing out this data can waste potentially important information. 
\cite{shiarlis2016inverse} investigates this topic in the Inverse Reinforcement Learning domain. 
It might be interesting to see a similar approach for Atari 2600 in Learning from Demonstration domain.

\footnotetext{\texttt{https://github.com/udacity/self-driving-car-sim}} 

The frameskip coefficient has been shown to be very important in RL~\cite{braylan2015frame,DBLP:journals/corr/SharmaLR17,DBLP:journals/corr/Lakshminarayanan16b}.  It would be interesting to investigate frameskip, using the human data that we can extract from the Atari Grand Challenge dataset.

Curriculum learning has proven useful in RL~\cite{leibfried2016deep}.
Having data of players with different expertise, we can investigate curriculum learning with respect to this.

We have also seen attempts to investigate how humans learn to play Atari~\cite{tsividis2017human}. Our dataset might be interesting for this kind of research.

In conclusion, we release a dataset of human Atari 2600 replays of five games, that is ${\sim}9.7$ Million frames or ${\sim}45$ hours of game play time.
We are describing its main properties, i.e. scale and diversity.
We show that in order to achieve high performance, it is more important to collect data of players with a high level of expertise, than to collect a lot of low-skilled data.
We plan to update the dataset in the future by adding ``professional'' Atari 2600 players data. 
We release the code for data collection as well, which gives the opportunity for everybody to extend the dataset.
We also show some of the possible research directions the Atari Grand Challenge dataset could be used in. 
We hope that our release will catalyze the research in sample-efficiency of RL and learning from human demonstration.

\subsubsection*{Acknowledgments}
Vitaly Kurin would like to thank Microsoft Research Cambridge for hosting him during the project and for the Microsoft Azure for Research grant. The authors would also like to thank Paulo Peccin, Javatari creator, for the emulator and useful discussions.
\bibliography{abbrev,references}
\bibliographystyle{abbrvnat}
\end{document}